\documentclass{article}

\usepackage[preprint]{neurips_2026}

\usepackage[utf8]{inputenc}
\usepackage[T1]{fontenc}
\usepackage{hyperref}
\usepackage{url}
\usepackage{booktabs}
\usepackage{amsfonts}
\usepackage{amsmath}
\usepackage{amssymb}
\usepackage{nicefrac}
\usepackage{microtype}
\usepackage[table]{xcolor}
\usepackage{graphicx}
\usepackage{wrapfig}
\usepackage{caption}
\usepackage{enumitem}
\usepackage{dsfont}
\usepackage{tikz}
\usetikzlibrary{positioning, fit, arrows.meta, calc}

\def\sysname{VACO}

\title{\sysname{}: Vertex-Aligned Decoding for Dense Hand Contact Estimation from a Single Image}

\author{%
  \normalfont Binhong Ye$^{1,*}$ \quad
  \normalfont Jin-Chuan Shi$^{1,*}$ \quad
  \normalfont Tao Liu$^{1}$ \quad
  \normalfont Xin Wang$^{1}$ \quad
  \normalfont Hao Chen$^{1,\dagger}$ \quad
  \normalfont Chunhua Shen$^{1,2,\dagger}$
  \\[0.5em]
  \normalfont $^{1}$State Key Lab of CAD \& CG, Zhejiang University, China
  \\
  \normalfont $^{2}$Zhejiang University of Technology, China
  \\[0.35em]
  \normalfont $^{*}$Equal contribution. $^{\dagger}$Corresponding author.
}

\date{}

\begin{document}

\maketitle

\begin{abstract}
Dense hand contact estimation from a single RGB image is a foundational primitive for dexterous manipulation and 3D interaction analysis. Direct predictors typically decode the entire per-vertex contact map from one shared query, leaving individual vertices ungrounded in the image and smearing contact across adjacent, non-touching surfaces. We present \textbf{\sysname{}}, a vertex-aligned decoder in which each MANO vertex samples local image features at its projected location. A hand-reconstruction backbone, jointly trained for reconstruction and contact, predicts these locations at inference without an auxiliary reconstructor. Training uses noise-perturbed ground-truth projections and end-to-end feature adaptation to improve robustness and contact-sensitive features. A single \sysname{} checkpoint sets a new state of the art on five benchmarks spanning hand--object, hand--human, hand--scene, and bimanual interactions. It improves F1 by $+20.7$ pp over the strongest prior direct predictor on Hi4D and generalizes to the held-out HOT3D and HOCap, with gains of $+5.5$ and $+2.7$\,pp. The resulting contact predictions also improve downstream HOI reconstruction.
\end{abstract}

\section{Introduction}
\label{sec:intro}

Dense hand contact maps serve as a foundational visual primitive for physical interaction systems. Dexterous robotic manipulation policies require precise knowledge of which finger surfaces will touch an object~\cite{Grady_2021_CVPR}, while immersive AR/VR and digital-human pipelines rely on dense contact to re-target plausible grasps. Although specialized hardware---such as instrumented gloves~\cite{Brahmbhatt_2019_CVPR}---can collect accurate contact labels in controlled laboratory settings, they cannot scale to unconstrained, in-the-wild images. Fortunately, dense hand contact is a property human observers can confidently infer from sight alone: given a still image of a hand interacting with a tool, one can easily distinguish which knuckles press the surface and which finger pads are lifted. This cognitive capability makes \emph{per-vertex contact prediction directly from a single RGB image}---estimating a binary label for every vertex of a parametric MANO~\cite{Romero_2017} hand---a desirable yet challenging visual task.

\begin{figure}[t]
    \centering
    \includegraphics[width=\linewidth]{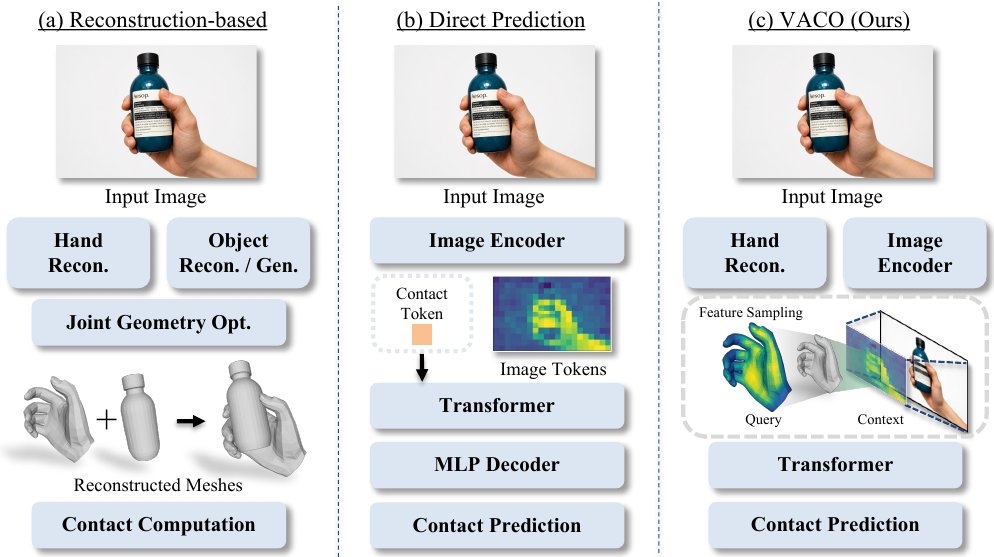}
    \caption{\textbf{Dense hand contact paradigms.} \textbf{(a)} Geometric proximity after reconstruction. \textbf{(b)} Shared-query direct prediction. \textbf{(c)} \sysname{} samples image evidence at projected vertices.}
    \label{fig:teaser}
\end{figure}

To date, two primary paradigms have tackled this image-to-contact problem, yet both leave critical cues unexploited. \emph{Reconstruction-based} approaches~\cite{Grady_2021_CVPR}, also called \emph{reconstruct-then-query}, first recover both the hand mesh and the interacting object, deriving contact from geometric proximity. While semantically explicit, this paradigm inherits the brittleness of full-object reconstruction, whose errors compound in cluttered scenes. Conversely, \emph{direct prediction} methods regress per-vertex contact straight from pixels~\cite{10.1007/978-3-031-19769-7_33,Hu_Zhang_Chen_Li_Wang_Liu_Sun_2024}. The strongest representative, HACO~\cite{jung2025learningdensehandcontact}, achieves state-of-the-art performance by addressing data imbalance through a vertex-level class-balanced loss. However, these fast and scalable predictors treat the entire vertex set as a flat, 1D topological target. By reading every per-vertex output through a \emph{single} shared contact query token---cross-attending to a spatial image feature grid and decoded by one shared linear head---the network has no per-vertex handle on the image evidence. The model never knows where on the image plane each vertex actually projects. This shared-query bottleneck creates a structural blind spot that produces \emph{geometric bleeding}: contact predictions spill from a touching vertex onto neighboring non-touching vertices on adjacent fingers and knuckles, even when only a single fingertip is actively touching.

We argue that the missing ingredient is a \emph{vertex-localized} representation. To mirror human visual reasoning, each contact decision should be made individually per vertex, directly consulting the image evidence under that vertex's own projected footprint. We operationalize this insight by introducing \textbf{\sysname{}} (\textbf{V}ertex-\textbf{A}ligned \textbf{CO}ntact), a decoding architecture that replaces the shared-query bottleneck with geometry-driven per-vertex image grounding. Modern hand pose estimators~\cite{Pavlakos_2024_CVPR,Potamias_2025_CVPR} already extract parametric meshes from unconstrained images in real time. Rather than querying recovered geometry for proximity, \sysname{} uses such a mesh purely as a \emph{spatial scaffold} for sampling: it tells each vertex where to look, and nothing more. As illustrated in Figure~\ref{fig:teaser}, we retain the speed of the direct prediction paradigm but replace its shared-query regression head with a vertex-aligned sampling mechanism.

Given an image, \sysname{} treats every MANO vertex as a query and samples multi-scale features at its 2D projection. Training uses noise-perturbed ground-truth projections, whereas inference uses projections from the co-trained reconstruction backbone. End-to-end adaptation makes the encoder sensitive to contact cues such as skin--object boundaries and occlusion edges.

Because \sysname{} relies on localized per-vertex signals, it benefits from consistent cross-dataset annotation. To isolate the contribution of our architecture from data curation, we train on the unified pool recently standardized by HACO~\cite{jung2025learningdensehandcontact}, which spans hand--object grasps, in-the-wild HOI, bimanual manipulation, and human--scene contact~\cite{Hasson_2019_CVPR,Chao_2021_CVPR,Fan_2023_CVPR,Liu_2022_CVPR,Shimada_2023_TOG} under a single proximity protocol. A single \sysname{} model trained on this pool generalizes across these interaction scenarios without dataset-specific heads.

Our contributions are threefold. (i) We introduce \textbf{\sysname{}}, a vertex-aligned decoder that samples image features at projected MANO vertices. Its co-trained hand-reconstruction backbone supplies these projections at inference, avoiding an auxiliary reconstructor. (ii) Controlled comparisons isolate the benefit of vertex-aligned decoding: with the same HaMeR backbone and without reconstruction co-training or projection noise, \sysname{} outperforms HACO; within \sysname{}, ablations identify mesh-aligned sampling and end-to-end encoder adaptation as the dominant contributors. (iii) A single \sysname{} checkpoint establishes new state-of-the-art results across five benchmarks spanning hand--object, hand--human, hand--scene, and bimanual interactions, including two held-out datasets; its contact predictions also improve downstream HOI reconstruction.

\section{Related Work}
\label{sec:related}

\textbf{Dense Hand Contact Estimation.}
Estimating physical contact from images is a fundamental yet challenging task due to severe mutual occlusions. \emph{Reconstruction-based} methods~\cite{Grady_2021_CVPR,Chen_2022_ECCV_AlignSDF,Fan_2024_CVPR_HOLD,Hu_Zhang_Chen_Li_Wang_Liu_Sun_2024} (also termed \emph{reconstruct-then-query}) explicitly recover both hand and object geometry and derive contact from physical proximity. However, they inherit the brittleness of full-object reconstruction, which frequently fails in unconstrained scenes. Conversely, \emph{direct prediction} approaches bypass object recovery to regress contact straight from pixels. The strongest representative, HACO~\cite{jung2025learningdensehandcontact}, addresses data imbalance with a vertex-level class-balanced loss. Yet, it relies on a single shared query token to decode the entire per-vertex contact map, structurally losing the fine-grained spatial grounding of individual vertices. Other parallel tracks explore body-scale contact~\cite{Hassan_2021_CVPR,Huang_2022_CVPR,Tripathi_2023_ICCV} or generative contact synthesis for grasping~\cite{Liu_2023_ICCV,nguyen2024clickdiff,10.1007/978-3-031-73390-1_17}, which diverge from our focus on perceptual hand-level accuracy. In contrast to HACO's global bottleneck, \sysname{} treats each MANO~\cite{Romero_2017} vertex as a distinct query, anchoring each contact decision to the localized image evidence under that vertex's projected footprint.

\textbf{Parametric Hand Reconstruction.}
Monocular hand reconstruction has advanced from early CNN regressors~\cite{Boukhayma_2019_CVPR,Ge_2019_CVPR} to robust Transformer-based architectures~\cite{Lin_2021_CVPR,Pavlakos_2024_CVPR,Potamias_2025_CVPR} and diffusion models~\cite{Ren_2023_ICCV}. While their primary objective is to recover MANO parameters, these networks establish a dense, reliable correspondence between the 3D topology and the 2D image plane. Rather than proposing a new pose estimator or deriving contact from recovered geometry by proximity, \sysname{} initializes from a state-of-the-art reconstructor (WiLoR~\cite{Potamias_2025_CVPR}) and co-trains it, so that a single network supplies both the \emph{spatial scaffold} and the pixel-aligned features for contact estimation. This lets us ground contact geometrically without ever deriving it from hand--object proximity.

\textbf{Pixel-Aligned 3D Decoding.}
Anchoring decoder tokens to localized image evidence is a proven paradigm in representation learning. In 2D recognition, DETR-family decoders~\cite{Carion_2020_ECCV,Zhu_2021_ICLR} sample local evidence using sparse, learned object queries. In 3D, pixel-aligned implicit functions (PIFu~\cite{Saito_2019_ICCV}, ECON~\cite{Xiu_2023_CVPR}) query continuous spatial points for clothed-human digitization, while DensePose~\cite{Guler_2018_CVPR} maps body pixels to a canonical UV chart. Closer to our setting, recent hand-centric works~\cite{lin2022pixelaligned,chen2025hort} project MANO vertices onto image features to iteratively refine mesh geometry. While prior works predominantly use localized sampling to regress \emph{geometry} or \emph{bounding boxes}, \sysname{} uniquely repurposes this mechanism for \emph{dense contact estimation}. By using a geometrically fixed set of mesh-aligned tokens instead of learned queries, we ground contact reasoning in pixel evidence, and let mesh-wide self-attention propagate physical correlations (e.g., bilateral finger closure).

\section{Method}
\label{sec:method}

\begin{figure*}[t]
    \centering
    \includegraphics[width=\linewidth]{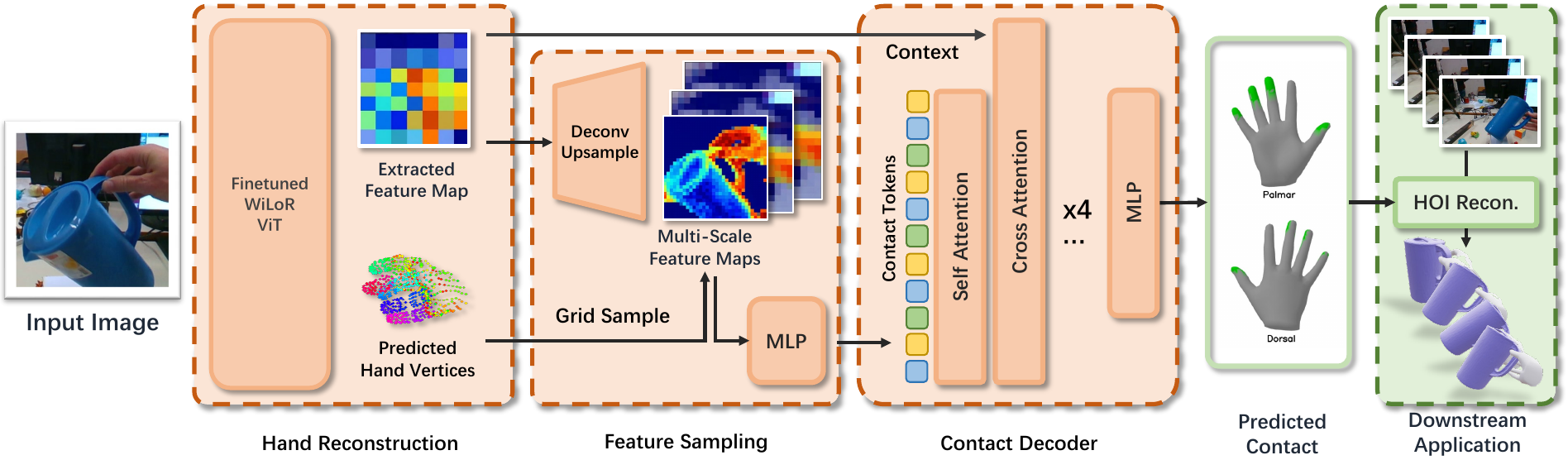}
    \caption{\textbf{\sysname{} pipeline.} A co-trained WiLoR~\cite{Potamias_2025_CVPR} backbone predicts MANO vertices and image features. Multi-scale sampling at vertex projections forms contact tokens; four decoder blocks combine mesh-wide self-attention with image cross-attention before per-vertex prediction.}
    \label{fig:pipe}
\end{figure*}

\subsection{Overview}
\label{sec:overview}

Given a single RGB image $I$, our goal is to predict a binary contact label $c_v \in \{0,1\}$ for every vertex $v \in \{1,\dots,V\}$ of the MANO~\cite{Romero_2017} mesh.

As illustrated in Figure~\ref{fig:pipe}, \textbf{\sysname{}} combines a WiLoR~\cite{Potamias_2025_CVPR} backbone with a geometry-driven contact decoder. The backbone encodes $I$ into a feature pyramid $\mathbf{F}_{\mathrm{vit}}$ and predicts a MANO mesh $\mathcal{V}$ that serves as the spatial scaffold. Projecting $\mathcal{V}$ onto the image yields vertex-aligned queries, which a lightweight Transformer decoder maps to contact probabilities:
\begin{equation}
\begin{aligned}
    I &\;\xrightarrow{\;\text{fine-tuned encoder}\;}\; \mathbf{F}_{\mathrm{vit}}, \\
    \big(\mathbf{F}_{\mathrm{vit}},\, \mathcal{V}\big) &\;\xrightarrow{\;\text{\sysname{} decoder}\;}\; \hat{\mathbf{c}} \in [0,1]^{V}.
\end{aligned}
\end{equation}
Predicted meshes can propagate pose errors into contact estimates. During training, we therefore sample at ground-truth projections $\pi(\mathcal{V}^{\mathrm{gt}})$ perturbed with image-plane Gaussian noise and jointly optimize the shared backbone for contact and reconstruction. At inference, the predicted scaffold replaces these training projections; Table~\ref{tab:scaffold} shows that it matches ground-truth projections in contact performance.

\subsection{Per-Vertex Token Construction}
\label{sec:vertex_token}

The central technical contribution of \sysname{} lies in formulating each MANO vertex as an independent query token that reads out the localized 2D visual evidence at its own projected footprint.

\textbf{Multi-scale visual sampling.} 
The raw ViT feature map, produced at the encoder's native patch stride, lacks the spatial acuity necessary for vertex-level discrimination. We first upsample $\mathbf{F}_{\mathrm{vit}}$ via a lightweight deconvolutional pyramid into $S$ progressively finer feature scales $\{\mathbf{F}^{s}_{\mathrm{vit}}\}_{s=1}^{S}$. For each scale $s$, we project the 3D vertices $\mathcal{V}$ onto the image plane using the weak-perspective camera parameters $\hat{\boldsymbol{\pi}}$ predicted by the same WiLoR head as $\mathcal{V}^{\mathrm{wilor}}$ (and from dataset annotation paired with $\mathcal{V}^{\mathrm{gt}}$ during training), and bilinearly sample the corresponding features:
\begin{equation}
\begin{aligned}
    \mathbf{f}^{\mathrm{vit}, s}_v &= \mathrm{Bilinear}\!\left(\mathbf{F}^s_{\mathrm{vit}},\; \pi(\mathbf{v}; \hat{\boldsymbol{\pi}})\right), \\
    \mathbf{f}^{\mathrm{vit}}_v &= \big[\mathbf{f}^{\mathrm{vit},1}_v,\, \dots,\, \mathbf{f}^{\mathrm{vit},S}_v\big].
\end{aligned}
\end{equation}
By fine-tuning the image encoder end-to-end with this explicit sampling grid, gradients flow directly from specific surface locations. This forces the multi-scale features to focus on localized contact boundaries rather than generic hand structures.

\subsection{Vertex-Aligned Decoding and Multi-Granularity Readout}
\label{sec:decoder}

The sampled feature for each of the $V$ vertices is projected into a $D$-dimensional token by a shallow MLP, and a learned per-vertex embedding $\mathbf{e}_v$, indexed by the fixed MANO vertex order, is added: $\mathbf{t}^{(0)}_v = \mathrm{MLP}(\mathbf{f}^{\mathrm{vit}}_v) + \mathbf{e}_v$. Tokens are kept in MANO vertex order, so each carries both its local appearance and the identity of the vertex it describes. These tokens are then fed into an $N$-layer Transformer decoder with $H$ attention heads and hidden dimension $D$. Hand-object interactions are structured; for instance, a power grasp involves strong correlations among all fingers, whereas a pinch relies on isolated fingertips. Each layer therefore performs two complementary reads. Self-attention runs globally over the set of $V$ vertex tokens, letting the decoder capture and propagate these structural contact patterns without hand-crafted heuristic rules. Cross-attention then queries the encoder's native-stride feature map $\mathbf{F}_{\mathrm{vit}}$, flattened into key/value tokens, so that a vertex can also condition on scene context beyond its own projected footprint---the interacting object, the opposing hand, or an occluder in front of it---which the point-wise sample at its projection cannot carry.

A shared contact readout head maps the refined vertex-token matrix
$\mathbf{T}\!\in\!\mathbb{R}^{V\times D}$ to per-vertex contact logits
$\hat{\boldsymbol{\ell}} = f_{\mathrm{out}}(\mathbf{T}) \in \mathbb{R}^{V}$, where
$f_{\mathrm{out}}$ is applied independently to each vertex token and outputs one scalar logit per vertex. The corresponding contact probabilities are given by
$\hat{\mathbf{c}} = \sigma(\hat{\boldsymbol{\ell}})$.
However, dense contact ground truth is derived from a proximity heuristic at the per-category threshold $\tau$ (following HACO's per-category protocol, $10$\,mm for most datasets; see App.~\ref{sec:supp_data}), making it noisy at the individual vertex level. Directly supervising this raw $V$-dimensional output leads to an unstable loss landscape. Following HACO, we adopt a coarse-to-fine multi-granularity readout that aggregates $\hat{\boldsymbol{\ell}}$ at four structural resolutions ($r \in \{V, V_2, V_3, J\}$, with $V > V_2 > V_3 > J$) through linear operators $\mathbf{R}_{r}\!\in\!\mathbb{R}^{n_r\times V}$: the qslim mesh-downsampling matrices of~\cite{Bouritsas_2019_ICCV} for $r\!\in\!\{V_2,V_3\}$, MANO's joint regressor~\cite{Romero_2017} for $r{=}J$, and the identity for $r{=}V$. The aggregated logits are
\begin{equation}
    \hat{\boldsymbol{\ell}}_{r} = \mathbf{R}_{r}\, \hat{\boldsymbol{\ell}}.
\end{equation}
This hierarchical design stabilizes optimization and yields joint- and part-level contact predictions as a byproduct, benefiting downstream manipulation tasks.

\subsection{Training Objective and Standardization}
\label{sec:loss_data}

A core challenge in dense contact estimation is data heterogeneity and extreme class imbalance (e.g., most vertices do not make contact). To keep the comparison with prior direct predictors as tight as possible on the data side, we follow the HACO framework~\cite{jung2025learningdensehandcontact} for dataset usage and adopt its VCB loss; the reconstruction term above is additional supervision that HACO does not use.

\textbf{Supervision strategy.} 
We optimize the network using the Vertex-Level Class-Balanced (VCB) loss $\mathcal{L}_{\mathrm{VCB}}$ proposed in~\cite{jung2025learningdensehandcontact}. For each granularity $r$, the VCB loss dynamically adjusts the Binary Cross-Entropy (BCE) penalty based on the historical contact frequency of each specific vertex, preventing the network from collapsing to trivial all-zero predictions. The total loss is the sum across all four granularities, augmented with an $L_1$ vertex-mean prior $\mathcal{L}_{\mathrm{reg}}$ and a mesh-adjacency smoothness term $\mathcal{L}_{\mathrm{smooth}}$ (defined in App.~\ref{sec:supp_impl}) to suppress isolated salt-and-pepper artifacts:
\begin{equation}
\begin{split}
    \mathcal{L}_{\mathrm{total}} ={}& \sum_{r\in\{V,V_2,V_3,J\}} \mathcal{L}^{r}_{\mathrm{VCB}} \\
    &+ \lambda_{\mathrm{reg}}\mathcal{L}_{\mathrm{reg}} + \lambda_{\mathrm{smooth}}\mathcal{L}_{\mathrm{smooth}} + \lambda_{\mathrm{rec}}\mathcal{L}_{\mathrm{rec}},
\end{split}
\end{equation}
The final term $\mathcal{L}_{\mathrm{rec}}$ is WiLoR's reconstruction objective, which preserves pose accuracy as the shared backbone specializes for contact. Projection perturbation further improves robustness to residual pose errors (Table~\ref{tab:ablation}, rows b--c).

\textbf{Data consistency.} 
Because vertex-aligned decoding relies on accurate hand-vertex correspondence, we adopt the unified dataset pool curated by HACO~\cite{jung2025learningdensehandcontact}. It standardizes DexYCB, HO3D, ARCTIC, Hi4D, and RICH under a common per-category contact protocol, ensuring consistent dense contact supervision across heterogeneous capture conditions. This unified representation enables joint training across lab-captured, in-the-wild, and human-scene interactions without dataset-specific tuning, allowing the model to learn generalizable contact patterns rather than dataset-dependent annotation biases.
\section{Implementation Details}
\label{sec:impl}

We implement \sysname{} in PyTorch~\cite{paszkepytorch}. The ViT-Huge image encoder is initialized from the official WiLoR~\cite{Potamias_2025_CVPR} checkpoint pre-trained on WHIM and fine-tuned jointly for contact prediction and hand reconstruction; its predicted mesh provides the inference scaffold. Training projections are perturbed with Gaussian noise of standard deviation $0.02$ of the crop side, truncated at $3\sigma$. We extract $S{=}3$ feature scales at strides $\{1{:}8,\,1{:}4,\,1{:}2\}$. The multi-granularity readout uses MANO resolutions $V{=}778$, $V_2{=}336$, $V_3{=}84$, and $J{=}21$, and the decoder comprises $N{=}4$ blocks with $H{=}8$ attention heads and hidden dimension $D{=}512$.

The total loss assigns unit weight to each of the four VCB terms and to the reconstruction objective ($\lambda_{\mathrm{rec}}{=}1.0$), with $\lambda_{\mathrm{reg}}{=}0.1$ for the vertex-mean prior and $\lambda_{\mathrm{smooth}}{=}1.0$ for mesh-adjacency smoothness. We jointly optimize the encoder and decoder with AdamW~\cite{loshchilov2018decoupled} (weight decay $10^{-4}$) at a base learning rate of $10^{-5}$; the encoder uses $0.5\times$ this rate, and both rates decay by $0.9\times$ at epoch~$5$. Training runs for $10$ epochs on $8$ NVIDIA RTX 4090 GPUs with effective batch size $64$ and FP16 mixed precision, taking approximately $4$ days ($\sim$768 GPU-hours). Following HaMeR/WiLoR~\cite{Pavlakos_2024_CVPR,Potamias_2025_CVPR}, we apply random scaling, rotation, color jitter, Gaussian noise, motion blur, and low-resolution degradation. At inference, per-vertex probabilities are thresholded at $\tau_{\mathrm{inf}}{=}0.52$ following HACO~\cite{jung2025learningdensehandcontact}; for left hands, we horizontally mirror the image and projected scaffold, then re-index predictions to the left-hand MANO topology. The reconstruction objective, per-dataset proximity protocol, and further implementation details are provided in Apps.~\ref{sec:supp_data} and~\ref{sec:supp_impl}.

\begin{figure*}[!t]
    \centering
    \includegraphics[width=\linewidth]{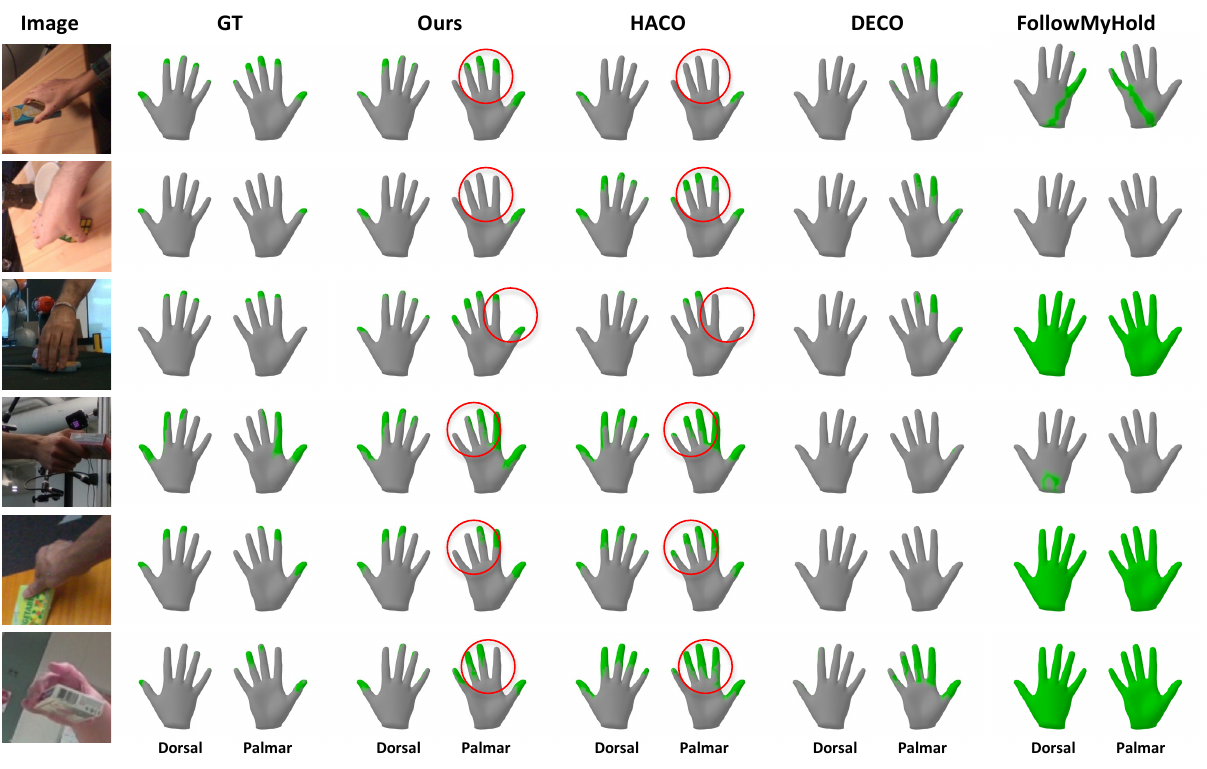}
    \caption{\textbf{Qualitative comparison.} Input, ground truth, and predictions from \sysname{}, HACO, DECO, and FollowMyHold. Green denotes contact; red circles mark corresponding regions across methods, exposing baseline smearing or omissions.}
    \label{fig:qual}
\end{figure*}

\section{Experiments}
\label{sec:exp}

We evaluate cross-domain generalization (Sec.~\ref{sec:exp_main}), analyze the model components and scaffold robustness (Sec.~\ref{sec:exp_abl}), and assess downstream HOI reconstruction (Sec.~\ref{sec:exp_downstream}).

\subsection{Experimental Setup}
\label{sec:exp_setup}

\textbf{Datasets and metrics.}
We compile a comprehensive training corpus covering five canonical interaction paradigms: hand--object grasping (ARCTIC~\cite{Fan_2023_CVPR}, DexYCB~\cite{Chao_2021_CVPR}, HO3D~\cite{Hampali_2020_CVPR}, H2O3D~\cite{Hampali_2022_CVPR}, HOI4D~\cite{Liu_2022_CVPR}, ObMan~\cite{Hasson_2019_CVPR}), bimanual manipulation (H2O~\cite{Kwon_2021_ICCV}), inter-hand/hand--human interaction (InterHand2.6M~\cite{Moon_2020_ECCV}, Hi4D~\cite{Yin_2023_CVPR}), hand--face interaction (Decaf~\cite{Shimada_2023_TOG}), and hand--scene contact (RICH~\cite{Huang_2022_CVPR}). Following HACO~\cite{jung2025learningdensehandcontact}, all eleven datasets are standardized under HACO's protocol, using vertex-to-surface proximity at a per-category threshold $\tau$ except where a dataset ships native per-vertex contact labels, yielding roughly $0.6$\,M training frames after frame subsampling following HACO.

For evaluation, we report results under two regimes. \emph{In-domain}: standard test splits of DexYCB (lab-based hand--object), Hi4D (close inter-hand/hand--human), and RICH (hand--scene), drawn from dataset families seen during training. \emph{Out-of-domain}: HOT3D~\cite{banerjee2024hot3d} (egocentric multi-view hand--object) and HOCap~\cite{wang2024hocap} (multi-view tabletop bimanual manipulation), both unseen during training. We report per-image Precision, Recall, and F1-Score on the binary per-vertex contact mask under HACO's per-category proximity protocol. All five columns of Table~\ref{tab:contact_main} use one checkpoint trained on the unified pool without per-benchmark fine-tuning; metrics are single-run point estimates given the $\sim$768 GPU-hour training cost (cf.\ App.~\ref{sec:supp_impl}). Inference uses the backbone's predicted scaffold.

\textbf{Baselines.}
We benchmark \sysname{} against three state-of-the-art methods spanning different contact-prediction paradigms: HACO~\cite{jung2025learningdensehandcontact}, the leading direct prediction model with a shared-query head; DECO~\cite{Tripathi_2023_ICCV}, a body-scale human--scene contact model retargeted to MANO; and FollowMyHold~\cite{aytekin2025followholdhandobjectinteraction}, a recent reconstruct-then-query method relying on per-image geometric optimization. 
\sysname{} is trained on the HACO-curated pool under the same contact protocol (Sec.~\ref{sec:loss_data}), and HACO is evaluated as released. Because \sysname{} additionally applies reconstruction supervision to its shared backbone, we report a matched HaMeR-backbone control (Table~\ref{tab:backbone}) whose \sysname{} variant omits both reconstruction co-training and projection noise. Due to the high computational cost of FollowMyHold ($\sim$9\,min/image), we evaluate it on a random 100-sample subset per test split, and mark it inapplicable on Hi4D and RICH, which lack the rigid hand-held object required for its proximity query.

\subsection{Main Results: Cross-Domain Generalization}
\label{sec:exp_main}

\begin{table*}[!t]
\centering
\caption{\textbf{Contact prediction across interaction paradigms.}
Per-image P/R/F1 under HACO's proximity protocol. The first three datasets are in-domain; $^{*}$ denotes held-out datasets. $^{\dagger}$FollowMyHold is evaluated on 100 samples per split at its reported thresholds (10\,mm for DexYCB/HOT3D; 5\,mm for HOCap); ``--'' denotes inapplicable settings. Best results are \textbf{bold}.}
\label{tab:contact_main}

\centering\footnotesize
\setlength{\tabcolsep}{2pt}
\begin{tabular}{l ccc ccc ccc ccc ccc}
\toprule
& \multicolumn{3}{c}{\textbf{DexYCB}} & \multicolumn{3}{c}{\textbf{Hi4D}} & \multicolumn{3}{c}{\textbf{RICH}} & \multicolumn{3}{c}{\textbf{HOT3D}$^{*}$} & \multicolumn{3}{c}{\textbf{HOCap}$^{*}$} \\
& \multicolumn{3}{c}{\small Hand--Object} & \multicolumn{3}{c}{\small Hand--Human} & \multicolumn{3}{c}{\small Hand--Scene} & \multicolumn{3}{c}{\small Hand--Object} & \multicolumn{3}{c}{\small Hand--Object} \\
\cmidrule(lr){2-4} \cmidrule(lr){5-7} \cmidrule(lr){8-10} \cmidrule(lr){11-13} \cmidrule(lr){14-16}
Method & P & R & F1 & P & R & F1 & P & R & F1 & P & R & F1 & P & R & F1 \\
\midrule
DECO                                              & 0.224 & 0.314 & 0.199 & 0.390 & 0.394 & 0.343 & 0.167 & 0.123 & 0.126 & 0.211 & 0.352 & 0.205 & 0.085 & 0.120 & 0.075 \\
FollowMyHold$^{\dagger}$   & 0.228 & 0.521 & 0.245 & --    & --    & --    & --    & --    & --    & 0.127 & 0.317 & 0.128 & 0.132 & 0.512 & 0.180 \\
HACO                                & 0.590 & 0.800 & 0.657 & 0.668 & 0.709 & 0.653 & 0.741 & 0.899 & 0.781 & 0.502 & \textbf{0.727} & 0.560 & 0.351 & 0.775 & 0.461 \\
\rowcolor{blue!5}
\textbf{\sysname{} (Ours)}                                                  & \textbf{0.606} & \textbf{0.859} & \textbf{0.693} & \textbf{0.811} & \textbf{0.943} & \textbf{0.860} & \textbf{0.818} & \textbf{0.908} & \textbf{0.840} & \textbf{0.579} & 0.723 & \textbf{0.615} & \textbf{0.367} & \textbf{0.842} & \textbf{0.488} \\
\bottomrule
\end{tabular}
\end{table*}

Table~\ref{tab:contact_main} details the quantitative results. \sysname{} establishes a new state of the art across all benchmarks, with its performance margins widening in complex interaction scenarios.

\textbf{In-domain generalization.}
On the three in-domain splits, \sysname{} improves over HACO by $+3.6$\,pp F1 on DexYCB, $+20.7$\,pp on Hi4D, and $+5.9$\,pp on RICH. The largest gain occurs on the highly occluded hand--human interactions of Hi4D, consistent with reduced cross-finger contact smearing. On RICH, \sysname{} improves both Precision ($+7.7$\,pp) and Recall ($+0.9$\,pp). DECO performs poorly across these splits, indicating that body-scale priors do not transfer readily to fine-grained hand contact.

\textbf{Out-of-domain robustness.}
Despite HOT3D and HOCap being unseen during training, \sysname{} outperforms HACO by $+5.5$ and $+2.7$\,pp F1, respectively, demonstrating robustness to novel camera viewpoints and bimanual grasps.

\textbf{Direct prediction vs.\ geometric optimization.}
FollowMyHold~\cite{aytekin2025followholdhandobjectinteraction}, representing the reconstruction-based paradigm, trails our feed-forward approach by a wide margin on its 100-sample evaluation subset (e.g., $-45$\,pp F1 on DexYCB, against our full-split score). Its relatively high Recall paired with exceptionally low Precision corroborates our hypothesis in Sec.~\ref{sec:related}: cascaded errors from object reconstruction propagate directly into the spatial proximity queries, yielding diffuse and inaccurate contact maps. \sysname{} bypasses these cascading errors entirely by grounding contact in visual features.

\textbf{Inference efficiency.}
On one RTX 4090, \sysname{} runs at $10$\,FPS (100\,ms/image), versus $21$\,FPS for DECO, $19$\,FPS for HACO, and approximately $9$\,min/image for FollowMyHold. Single-scale sampling removes the multi-scale overhead but costs $1.6$\,pp mean F1 (Table~\ref{tab:ablation}).

\textbf{Qualitative results.}
Figure~\ref{fig:qual} complements the quantitative comparison: HACO often spreads contact across neighboring fingers and the palm, whereas \sysname{} localizes it to the contacting phalanges, particularly in fine-grained interactions such as pinching.

\subsection{Ablations and Scaffold Analysis}
\label{sec:exp_abl}

\begin{table}[t]

\centering
\caption{\textbf{Ablations averaged over five benchmarks.} Rows (b)--(c) are cumulative; rows (d)--(g) independently modify the full model (a). $^{\ddagger}$Row (e) replaces the sampled features with random features at test time without retraining; all other variants are trained. Text deltas use unrounded values.}
\label{tab:ablation}
\footnotesize
\setlength{\tabcolsep}{4pt}
\begin{tabular}{c l ccc}
    \toprule
        & Variant
        & P $\uparrow$ & R $\uparrow$ & F1 $\uparrow$ \\
    \midrule
    \rowcolor{blue!5}
    (a) & Full \sysname{} (Ours)                          & \textbf{0.636} & \textbf{0.855} & \textbf{0.699} \\
    \midrule
    \multicolumn{5}{l}{\emph{training recipe}} \\
    (b) & w/o reconstruction co-training                  & 0.616 & 0.846 & 0.685 \\
    (c) & \quad and w/o projection noise                  & 0.618 & 0.844 & 0.684 \\
    \midrule
    \multicolumn{5}{l}{\emph{model and features}} \\
    (d) & w/o encoder fine-tuning                         & 0.481 & 0.688 & 0.486 \\
    (e) & w/o mesh-aligned sampling$^{\ddagger}$            & 0.500 & 0.657 & 0.471 \\
    (f) & single-scale sampling ($S{=}1$)      & 0.630 & 0.822 & 0.684 \\
    (g) & decoder self-attn.\ $\rightarrow$ MLP  & 0.605 & 0.851 & 0.678 \\
    \bottomrule
\end{tabular}
\end{table}

We ablate the scaffold training recipe and decoding head on the five benchmarks of Table~\ref{tab:contact_main}, with results in Table~\ref{tab:ablation}. Rows (b)--(d) and (f)--(g) are trained with the same data pool and $10$-epoch schedule; rows (b)--(c) vary the training objective and projection augmentation, and row (e) is a test-time intervention described below.

\textbf{Training recipe.}
Rows (b) and (c) isolate reconstruction co-training and projection noise. Without reconstruction supervision (row b), the predicted scaffold drifts from the geometry used to train the decoder, reducing F1 by $1.4$\,pp. Removing projection noise as well (row c) costs a further $0.2$\,pp, showing that reconstruction co-training drives most of this gain.

\textbf{Encoder fine-tuning.}
Freezing the ViT encoder (row d) restricts the decoder to features optimized solely for geometric reconstruction, leading to a noticeable drop in F1. End-to-end fine-tuning allows the multi-scale feature pyramid to adapt to hyper-local visual cues (e.g., skin indentations and boundaries) critical for contact estimation.

\textbf{Mesh-aligned sampling.}
Row (e) replaces the sampled features by random features at test time without retraining; F1 falls to $0.471$. Freezing the encoder (row d) has a comparable effect. These components are jointly load-bearing: row (e) is disproportionately affected by a recall collapse on contact-dense Hi4D ($R{=}0.098$), and excluding Hi4D makes it the milder ablation.

\textbf{Multi-scale sampling and decoder self-attention.}
Rows (f) and (g) probe the two design choices \emph{inside} the vertex-aligned pathway, and both turn out to be refinements rather than load-bearing parts. Sampling a single scale instead of the $S{=}3$ pyramid (row f) costs $1.6$\,pp F1 and shifts the operating point, trading recall for precision. Replacing the decoder's self-attention over the $V$ vertex tokens with a per-token MLP (row g), while retaining the cross-attention to the context features, costs $2.1$\,pp. Set against the $21$\,pp that row (d) moves and the $23$\,pp that the test-time row (e) costs, both are refinements on top of per-vertex grounding rather than sources of the bulk of the gain; we do not claim they are inert, since all entries are single-run point estimates.

\begin{figure*}[t]
    \centering
    \includegraphics[width=\linewidth]{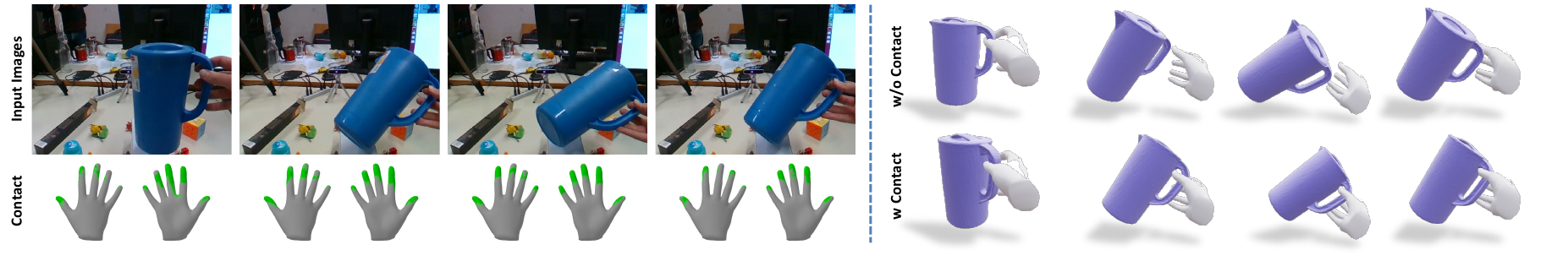}
    \caption{\textbf{Qualitative HOI reconstruction.} Left: input frames and \sysname{} contacts (green). Right: AGILE without and with the contact prior, each shown from four viewpoints. The prior yields a tighter grasp with fewer floating-finger and interpenetration artifacts.}
    \label{fig:downstream_agile}
\end{figure*}

\begin{table}[t]
\centering
\caption{\textbf{Matched HaMeR-backbone comparison over five benchmarks.} Both encoders are fine-tuned with the same initialization, data pool, augmentations, losses, optimizer and schedule; HACO uses its released weights. This \sysname{} variant omits reconstruction co-training and projection noise, and takes its sampling locations from official WiLoR predictions.}
\label{tab:backbone}
\footnotesize
\setlength{\tabcolsep}{4pt}
\begin{tabular}{lccc}
    \toprule
    Method (HaMeR backbone) & P $\uparrow$ & R $\uparrow$ & F1 $\uparrow$ \\
    \midrule
    HACO~\cite{jung2025learningdensehandcontact} & 0.570 & 0.782 & 0.622 \\
    \rowcolor{blue!5}
    \sysname{} (Ours) & \textbf{0.619} & \textbf{0.825} & \textbf{0.677} \\
    \bottomrule
\end{tabular}
\end{table}

\textbf{Matched-backbone comparison.}
To control for backbone choice, we retrain \sysname{} with HACO's HaMeR image encoder at $256{\times}256$, matching the initialization, data pool, augmentations, contact losses, optimizer and learning-rate schedule, and training for the same $10$ epochs; reconstruction co-training and projection noise are omitted, and the sampling locations come from official WiLoR predictions while HACO is evaluated with its released weights. The contact decoder and the source of the sampling scaffold are therefore the principal differences from HACO in this control, which keeps the VCB and regularization losses identical. \sysname{} improves all three metrics, including mean F1 from $0.622$ to $0.677$ (Table~\ref{tab:backbone}), supporting attribution of the gain to the vertex-aligned decoder rather than to the feature extractor alone.

\begin{table}[t]

\centering
\caption{\textbf{Inference scaffold quality over five benchmarks.} Only 2D sampling locations vary. The noise $\sigma$, measured relative to crop size, is $1\times$, $2.5\times$, or $5\times$ the training level.}
\label{tab:scaffold}
\footnotesize
\setlength{\tabcolsep}{4pt}
\begin{tabular}{lccc}
    \toprule
    Vertices projected at inference             & P $\uparrow$ & R $\uparrow$ & F1 $\uparrow$ \\
    \midrule
    \rowcolor{blue!5}
    Our own prediction (default)                & 0.636 & 0.855 & \textbf{0.699} \\
    Ground-truth projections (oracle)           & 0.630 & \textbf{0.866} & \textbf{0.699} \\
    \midrule
    \multicolumn{4}{l}{\emph{ground-truth projections perturbed at inference}} \\
    \quad $\sigma{=}0.02$                       & \textbf{0.637} & 0.849 & 0.697 \\
    \quad $\sigma{=}0.05$                       & 0.634 & 0.838 & 0.691 \\
    \quad $\sigma{=}0.10$                       & 0.625 & 0.802 & 0.673 \\
    \bottomrule
\end{tabular}
\end{table}

\begin{table}[t]

\centering
\caption{\textbf{Hand reconstruction over five benchmarks.} We compare the co-trained \sysname{} backbone with its WiLoR~\cite{Potamias_2025_CVPR} initialization; the last block separates evaluation regimes.}
\label{tab:recon}
\footnotesize
\setlength{\tabcolsep}{4pt}
\begin{tabular}{lcc}
    \toprule
    Metric & WiLoR & \sysname{} \\
    \midrule
    AUC-J $\uparrow$              & 0.839 & \textbf{0.870} \\
    PA-MPJPE (mm) $\downarrow$    & 8.05  & \textbf{6.52} \\
    AUC-V $\uparrow$              & 0.846 & \textbf{0.866} \\
    PA-MPVPE (mm) $\downarrow$    & 7.72  & \textbf{6.70} \\
    F@5 $\uparrow$                & 0.642 & \textbf{0.704} \\
    F@15 $\uparrow$               & 0.977 & \textbf{0.983} \\
    \midrule
    \multicolumn{3}{l}{\emph{PA-MPJPE (mm), by evaluation regime}} \\
    \quad in-domain (3 splits)    & 8.45  & \textbf{5.52} \\
    \quad held-out (2 splits)     & \textbf{7.45} & 8.00 \\
    \bottomrule
\end{tabular}
\end{table}

\textbf{Scaffold robustness.}
Table~\ref{tab:scaffold} varies only the 2D sampling locations while holding the trained model fixed. The predicted scaffold matches the oracle's mean F1 ($0.699$) and is $0.4$\,pp better in-domain but $0.6$\,pp worse on held-out splits. Moderate projection errors are well tolerated: relative to oracle projections, F1 drops by $0.2$\,pp at the training noise level and $0.8$\,pp at $2.5\times$ that level, reaching $0.673$ only at $5\times$. The predicted scaffold even outperforms the oracle perturbed at the training noise level ($0.699$ vs.\ $0.697$). Since it already matches the unperturbed oracle on average, an external reconstructor offers no aggregate benefit under this evaluation.

Because the backbone predicts the scaffold it consumes, reconstruction accuracy is part of the method rather than a property of an external module. Relative to its WiLoR initialization, co-training improves every aggregate metric in Table~\ref{tab:recon}. In-domain PA-MPJPE improves from $8.45$ to $5.52$\,mm, whereas held-out PA-MPJPE increases from $7.45$ to $8.00$\,mm; the reported held-out contact results already include this weaker scaffold. Co-training improves in-domain reconstruction, with only a modest trade-off observed on held-out data.

\subsection{Downstream Application: Physically Grounded HOI Reconstruction}
\label{sec:exp_downstream}

\begin{table}[t]

\centering
\caption{\textbf{HOI reconstruction on HO3D-v3.} Right-hand CD (mm) and F-scores (\%), reported as mean$_{\pm\sigma}$ across sequences.}
\label{tab:downstream_agile}
\footnotesize
\setlength{\tabcolsep}{3pt}
\begin{tabular}{lccc}
\toprule
                               & CD $\downarrow$ & F@5 $\uparrow$ & F@10 $\uparrow$ \\
\midrule
AGILE                          & $23.11_{\pm 22.77}$ & $12.07_{\pm 8.15}$ & $30.43_{\pm 16.75}$ \\
\rowcolor{blue!5}
+\,\sysname{} contact          & $\mathbf{12.46}_{\pm 8.65}$ & $\mathbf{15.32}_{\pm 4.79}$ & $\mathbf{37.87}_{\pm 11.15}$ \\
\bottomrule
\end{tabular}
\end{table}

Dense contact masks can provide geometric priors for resolving depth ambiguity and enforcing physical plausibility in 3D reconstruction. We integrate \sysname{} into AGILE~\cite{shi2026agile} as a frozen, image-conditioned contact prior: an SDF-based contact loss pulls the predicted-contact vertices toward the object surface from both sides, whereas vertices outside the predicted contact set do not enter this term. All other AGILE losses remain unchanged; the contact-mask threshold and the gating rule are specified in App.~\ref{sec:supp_downstream}. On four HO3D-v3 test sequences held out from training (AP14, MPM13, SB11, SM1), Table~\ref{tab:downstream_agile} shows that the contact prior nearly halves right-hand Chamfer distance and improves both F-scores under otherwise identical settings; only AP14 additionally involves an object instance unseen during training, whereas the other three objects do appear in training data. The corresponding reconstructions are visualized in Figure~\ref{fig:downstream_agile}.

\section{Conclusion}
\label{sec:conclusion}

We introduced \sysname{}, grounding dense contact predictions in image features sampled at projected MANO vertices. Joint reconstruction and contact training lets the backbone predict these locations at inference. One checkpoint sets a new state of the art across hand--object, hand--human, hand--scene, and bimanual benchmarks and improves downstream HOI reconstruction with AGILE.

Two limits are worth naming. Accuracy remains bounded by the predicted scaffold: under heavy occlusion or truncation, pose and contact errors are coupled, and the perturbation sweep of Table~\ref{tab:scaffold} covers only the measured regime. Grounding is also confined to the image plane: vertices whose projections coincide---most notably palmar and dorsal surfaces---sample identical image features. The learned per-vertex embedding lets such vertices produce different outputs, but the decoder receives no explicit depth, normal or visibility cue, so strongly overlapping surfaces remain ambiguous. The per-frame formulation omits temporal and cross-mesh reasoning, while proximity-derived labels limit supervision; temporal, multi-mesh, and geometric cues are natural extensions.
These limitations also suggest promising future directions. Temporal reasoning, multi-mesh interaction modeling, and geometric cues may further extend VACO toward more robust physical interaction understanding. More broadly, vertex-aligned decoding provides a general paradigm for grounding dense predictions in visual evidence.

\bibliographystyle{plainnat}
\bibliography{ref}

\clearpage
\appendix
\section{Datasets and Training Pool}
\label{sec:supp_data}

We train \sysname{} on a joint pool of public hand-interaction datasets curated by HACO~\cite{jung2025learningdensehandcontact}. For datasets without explicit contact annotations, we follow HACO and derive per-vertex labels by computing each MANO vertex's nearest-point distance to the interacting object or scene mesh and thresholding it at a per-category proximity $\tau$. For Hi4D~\cite{Yin_2023_CVPR} and Decaf~\cite{Shimada_2023_TOG}, which already release per-vertex contact ground truth, we use the released annotations directly. Table~\ref{tab:supp_datasets} lists the 11 datasets, their interaction domains, the post-subsampling frame counts, and the contact source. Counts are taken after HACO's frame subsampling; the per-epoch counts after balanced contact-difficulty sampling vary by run and are logged at training launch.

\section{Additional Implementation Details}
\label{sec:supp_impl}

This section expands on the training recipe summarized in the main paper.

\textbf{Optimization.} The ViT-Huge image encoder is initialized from the official WiLoR~\cite{Potamias_2025_CVPR} checkpoint pre-trained on WHIM. Encoder and decoder are fine-tuned jointly with AdamW~\cite{loshchilov2018decoupled} (weight decay $10^{-4}$) at a base learning rate of $10^{-5}$, with the encoder using $0.5\times$ the decoder rate to preserve its geometric priors and a $0.9\times$ decay applied at epoch~$5$, for $10$ epochs on $8$ NVIDIA RTX 4090 GPUs (effective batch $64$, FP16 mixed precision).

\textbf{Loss weights.} The total loss of the main paper sums four VCB terms (one per granularity, unit weight) with an $L_1$ vertex-mean prior ($\lambda_{\mathrm{reg}}{=}0.1$), a mesh-adjacency smoothness term ($\lambda_{\mathrm{smooth}}{=}1.0$), and the reconstruction objective at unit weight ($\lambda_{\mathrm{rec}}{=}1.0$).

\textbf{Reconstruction objective.} $\mathcal{L}_{\mathrm{rec}}$ is the objective WiLoR~\cite{Potamias_2025_CVPR} uses to train its hand reconstructor, which supervises the regressed mesh and MANO parameters wherever the annotation is available:
\begin{equation}
\begin{aligned}
    \mathcal{L}_{\mathrm{rec}} ={}& w_{\mathrm{k3d}}\|\mathbf{J}_{3\mathrm{D}}-\hat{\mathbf{J}}_{3\mathrm{D}}\|_1 + w_{\mathrm{k2d}}\|\pi(\mathbf{J}_{3\mathrm{D}})-\hat{\mathbf{J}}_{2\mathrm{D}}\|_1 \\
    &+ w_{\mathrm{v3d}}\|\mathbf{V}_{3\mathrm{D}}-\hat{\mathbf{V}}_{3\mathrm{D}}\|_1 \\
    &+ w_{\mathrm{v2d}}\|\pi(\mathbf{V}_{3\mathrm{D}})-\hat{\mathbf{V}}_{2\mathrm{D}}\|_1 \\
    &+ w_{\mathrm{go}}\|\boldsymbol{\theta}_{\mathrm{go}}-\hat{\boldsymbol{\theta}}_{\mathrm{go}}\|_2^2 \\
    &+ w_{\mathrm{hp}}\|\boldsymbol{\theta}_{\mathrm{hp}}-\hat{\boldsymbol{\theta}}_{\mathrm{hp}}\|_2^2 + w_{\beta}\|\boldsymbol{\beta}-\hat{\boldsymbol{\beta}}\|_2^2,
\end{aligned}
\label{eq:rec}
\end{equation}
where $\pi(\cdot)$ is the weak-perspective projection, $\mathbf{J}$ and $\mathbf{V}$ are MANO joints and vertices, and $\boldsymbol{\theta}_{\mathrm{go}}$, $\boldsymbol{\theta}_{\mathrm{hp}}$, $\boldsymbol{\beta}$ are the global orientation, hand articulation and shape parameters. Every term weight is left at WiLoR's released value: $w_{\mathrm{k3d}}{=}0.05$, $w_{\mathrm{k2d}}{=}0.01$, $w_{\mathrm{v3d}}{=}w_{\mathrm{v2d}}{=}0$, $w_{\mathrm{go}}{=}w_{\mathrm{hp}}{=}0.001$, and $w_{\beta}{=}0.0005$; we tune none of them. Note that the two vertex terms carry zero weight in that configuration, so the geometric supervision reaches the shared backbone through the joint keypoints and the MANO parameters rather than through dense vertex correspondences. The only component of WiLoR's objective we do not use is its adversarial term, the discriminator that penalizes implausible articulations: we leave it disabled, so the reconstruction branch is supervised purely by the annotated targets in Eq.~\ref{eq:rec}.

\textbf{Augmentations.} Following HaMeR/WiLoR~\cite{Pavlakos_2024_CVPR,Potamias_2025_CVPR}, the cropped image receives random scaling, rotation, color jittering, Gaussian noise, motion blur, and low-resolution degradation.

\textbf{Regularization losses.} The two auxiliary terms in the main paper's total loss act on the per-vertex probabilities $\hat{\mathbf{c}}=\sigma(\hat{\boldsymbol{\ell}})$. The vertex-mean prior anchors the predicted contact rate to the ground-truth contact rate per frame,
\begin{equation}
    \mathcal{L}_{\mathrm{reg}} = \Big|\,\tfrac{1}{V}\!\sum_{v=1}^{V} \hat{c}_v \;-\; \tfrac{1}{V}\!\sum_{v=1}^{V} c^{\mathrm{gt}}_v\,\Big|,
\end{equation}
which prevents the network from drifting toward all-zero or all-one predictions. The mesh-adjacency smoothness term penalizes high-frequency disagreement along the MANO mesh edges $\mathcal{E}$,
\begin{equation}
    \mathcal{L}_{\mathrm{smooth}} = \tfrac{1}{|\mathcal{E}|}\!\sum_{(u,v)\in\mathcal{E}}\!\big(\hat{c}_u - \hat{c}_v\big)^{2},
\end{equation}
suppressing isolated salt-and-pepper activations without blurring true contact boundaries.

\begin{table*}[t]
\centering
\caption{\textbf{Training datasets used by \sysname{}.} ``Domain'' summarizes the interaction type and the number of hands annotated per frame. ``\# Frames'' lists the number of frames used per dataset after HACO's frame subsampling. ``$\tau$'' is the per-category proximity threshold (mm) used to derive per-vertex contact labels following HACO; ``native'' indicates that the dataset ships its own per-vertex contact ground truth.}
\label{tab:supp_datasets}

\small
\setlength{\tabcolsep}{6pt}
\begin{tabular}{llrl}
\toprule
Dataset & Domain & \# Frames & $\tau$ \\
\midrule
ARCTIC~\cite{Fan_2023_CVPR}             & bimanual hand--object       & 147{,}235 & 10\,mm \\
Decaf~\cite{Shimada_2023_TOG}           & single-hand hand--face      & 65{,}310  & native \\
DexYCB~\cite{Chao_2021_CVPR}            & single-hand hand--object    & 29{,}640  & 10\,mm \\
H2O~\cite{Kwon_2021_ICCV}               & bimanual hand--object       & 33{,}465  & 10\,mm \\
H2O3D~\cite{Hampali_2022_CVPR}          & bimanual hand--object       & 54{,}840  & 10\,mm \\
Hi4D~\cite{Yin_2023_CVPR}               & two-person, four hands      & 24{,}315  & native \\
HO3D (v2)~\cite{Hampali_2020_CVPR}      & bimanual hand--object       & 62{,}775  & 10\,mm \\
HOI4D~\cite{Liu_2022_CVPR}              & bimanual hand--object       & 84{,}425  & 10\,mm \\
InterHand2.6M~\cite{Moon_2020_ECCV}     & bimanual hand--hand         & 22{,}515  & 5\,mm  \\
ObMan~\cite{Hasson_2019_CVPR}           & single-hand (synthetic)     & 85{,}685  & 10\,mm \\
RICH~\cite{Huang_2022_CVPR}             & bimanual hand--scene        & 15{,}730  & 35\,mm \\
\midrule
\textbf{Total}                          &                             & \textbf{625{,}935} & \\
\bottomrule
\end{tabular}
\end{table*}

\textbf{Compute.} Training is run under DistributedDataParallel and takes approximately $4$ days ($\sim 768$ GPU-hours).

\section{Integrating \sysname{} Contact Priors into AGILE}
\label{sec:supp_downstream}

This section details how \sysname{}'s per-vertex contact predictions are consumed by AGILE~\cite{shi2026agile}, an optimization-based hand-object reconstruction framework, in the downstream evaluation reported in the main paper. The integration is engineered so that \sysname{} acts as an auxiliary contact prior alongside AGILE's existing geometric losses; only its interaction gate is affected, as detailed below.

\textbf{Where the prior plugs in.} AGILE optimizes a posed MANO mesh and an object pose against multi-view evidence in two stages: (i) a hand-only refinement stage that tightens the hand against image masks and predicted 2D hand joints, followed by (ii) a hand--object alignment stage that jointly refines both poses while encouraging plausible interaction. We supply \sysname{} contact predictions as a frozen, image-conditioned signal that gates an SDF-based contact loss in stage (ii); stage (i) is left unchanged.

\textbf{Contact mask from \sysname{} probabilities.} Given the per-vertex contact probabilities $\hat{\mathbf{c}} \in [0,1]^{V}$ produced by \sysname{} for the current frame, we form a binary mask
\begin{equation}
    M_v = \mathbf{1}\!\left[\hat{c}_v \geq \tau_{\mathrm{prob}}\right], \qquad
    \mathcal{C} = \{\, v : M_v = 1 \,\},
\end{equation}
and only enable the contact loss when $|\mathcal{C}| \geq \tau_{\mathrm{count}}$. We set $\tau_{\mathrm{prob}}=0.52$ (matching the inference threshold used in the main paper) and $\tau_{\mathrm{count}}=100$ vertices; the count gate suppresses spurious activations on near-empty masks.

\textbf{SDF-based contact loss.} Let $\Phi(\cdot)$ denote the signed distance field of the optimized object mesh, evaluated in the object-local frame and clipped to $|\Phi| < 1$\,m. Only the predicted contact set $\mathcal{C}$ is supervised, through a two-sided surface constraint that pulls those vertices onto the object surface whether they lie outside ($\Phi > 0$) or inside ($\Phi < 0$) the object:
\begin{equation}
\begin{aligned}[b]
    \mathcal{L}_{\mathrm{sdf\text{-}contact}}
    = \frac{1}{|\mathcal{C}|}\sum_{i \in \mathcal{C}} \Big[ &\lambda_{\mathrm{ext}}\max\!\big(0,\, \Phi(\mathbf{v}_i)\big) \\
    &+ \lambda_{\mathrm{int}}\max\!\big(0,\, -\Phi(\mathbf{v}_i)\big) \Big],
\end{aligned}
    \label{eq:sdf_contact}
\end{equation}
with $\lambda_{\mathrm{ext}}=5.0$ and $\lambda_{\mathrm{int}}=3.0$. Vertices outside $\mathcal{C}$ do not enter this term. Compared to AGILE's native interaction term, which weights every vertex by a soft SDF-distance gate, Eq.~\ref{eq:sdf_contact} replaces this gate with the explicit, image-conditioned binary mask $\mathcal{C}$, focusing the supervision on the surfaces \sysname{} predicts as contacting; it is a soft surface-attraction constraint rather than a guarantee that the reconstructed hand never penetrates the object.

\textbf{Two-stage schedule.} The contact loss is disabled while AGILE initializes the object pose, then activated together with stage~(ii) using the same $\lambda_{\mathrm{ext}}/\lambda_{\mathrm{int}}$ values listed above. All other AGILE losses (silhouette, semantic, regularization, temporal) retain the weights from the original framework.

\textbf{Implementation notes.} \sysname{} runs once per frame and predicts its own scaffold; because AGILE initializes from a WiLoR mesh of the same MANO topology, the per-vertex probabilities, the binary mask, and the optimized vertices are in correspondence without any resampling. Because $\hat{\mathbf{c}}$ does not depend on the optimization variables, the SDF contact loss adds only one signed-distance lookup per vertex per iteration.

\section{Additional Qualitative Comparisons}
\label{sec:supp_qual}

\begin{figure*}[t]
    \centering
    \includegraphics[width=\linewidth]{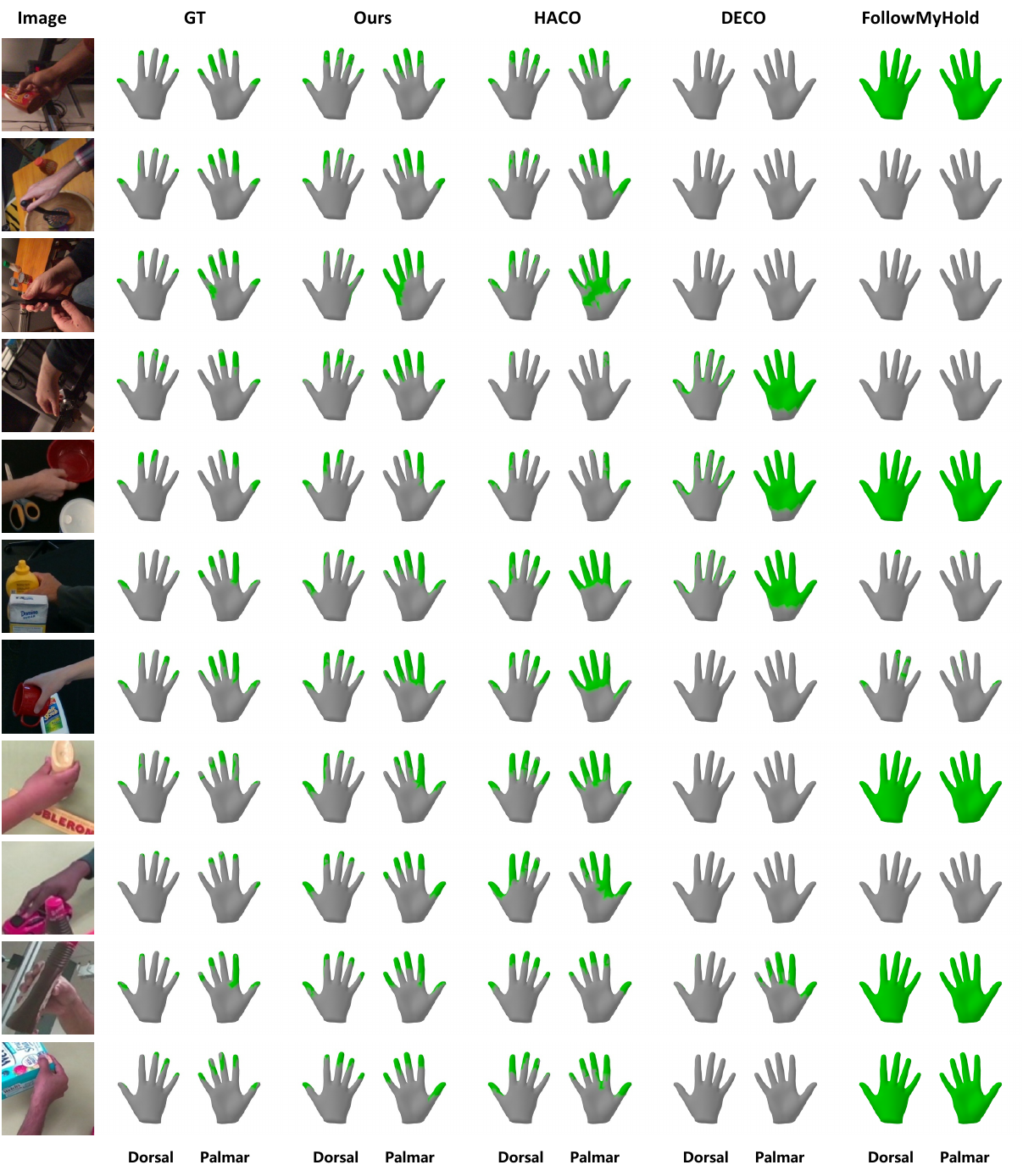}
    \caption{\textbf{Additional qualitative comparisons on DexYCB, HOCap, and HOT3D.} For each row we show the input image, the ground-truth contact map, and per-vertex contact predictions from HACO~\cite{jung2025learningdensehandcontact}, DECO~\cite{Tripathi_2023_ICCV}, and \sysname{} (Ours). \sysname{} consistently produces sharper, vertex-localized contact patterns, while shared-query baselines smear activations across adjacent fingers.}
    \label{fig:supp_comp}
\end{figure*}

Figure~\ref{fig:supp_comp} provides extra qualitative comparisons that complement the main paper's qualitative panel. We sample frames from three benchmarks that span distinct interaction regimes: DexYCB~\cite{Chao_2021_CVPR} (single-hand object grasps in a controlled lab capture), HOCap~\cite{wang2024hocap} (longer multi-view tabletop manipulation sequences with diverse object geometries), and HOT3D~\cite{banerjee2024hot3d} (egocentric headset captures with frequent self-occlusion). For each frame we visualize the input image, the ground-truth contact map, and the per-vertex predictions of \sysname{} alongside HACO and DECO. Across all three datasets, \sysname{} produces sharper contact boundaries that follow the actual finger surfaces touching the object, whereas the shared-query baselines smear contact across neighboring fingers and knuckles. The HOT3D rows additionally illustrate \sysname{}'s robustness under egocentric self-occlusion, where parts of the hand are truncated by the wrist or another finger.

\section{Left-Hand and Multi-Hand Inference}
\label{sec:supp_lefthand}

\begin{figure*}[t]
    \centering
    \includegraphics[width=\linewidth]{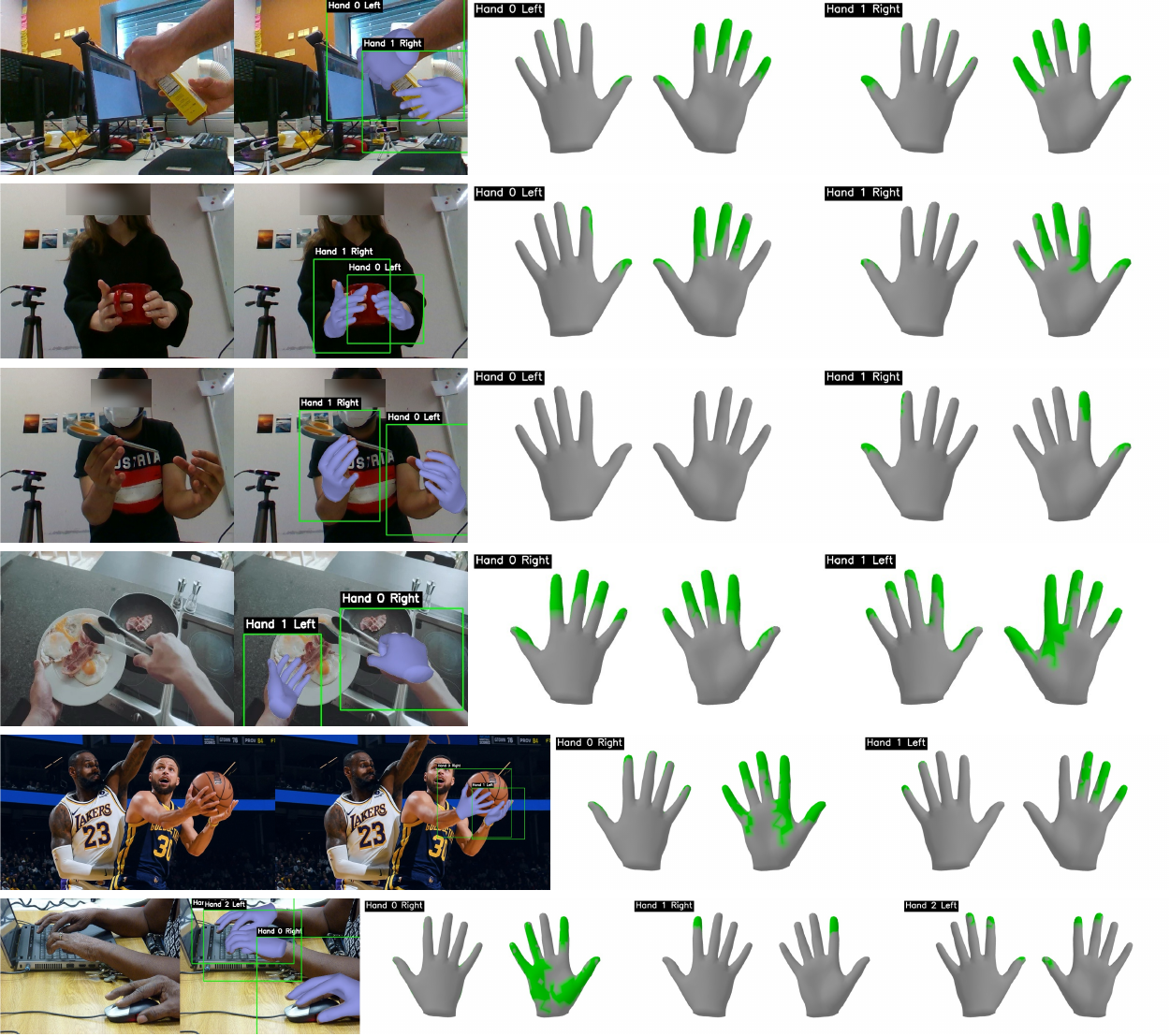}
    \caption{\textbf{Multi-hand contact predictions on in-the-wild scenes.} Each example contains 2--3 hands of mixed handedness; \sysname{} is invoked once per detected hand, with left hands routed through the mirroring trick and right hands run natively. Predictions remain vertex-localized and consistent across hands within the same scene.}
    \label{fig:supp_lefthand}
\end{figure*}

\sysname{} is trained exclusively on right hands, yet the model is readily applied to left hands at inference through a symmetric mirroring trick. Given an input frame whose handedness flag (provided by the upstream hand detector) indicates a left hand, we horizontally flip the image and the projected scaffold mesh so that the left hand appears as a right hand to the network. \sysname{} then runs unchanged on this mirrored input, and the predicted per-vertex contact probabilities are mapped back to the original left-hand MANO topology through the symmetric MANO vertex correspondence. Because MANO's left and right templates differ only by this fixed permutation, the trick adds no extra training, no architectural change, and only one image flip plus one vertex re-indexing per frame at inference. The same procedure scales to multi-hand scenes: each detected hand is processed independently, with right hands run natively and left hands routed through the mirroring trick.

Figure~\ref{fig:supp_lefthand} demonstrates this procedure on in-the-wild images that contain multiple hands per scene. Predictions remain vertex-localized for both left and right hands and capture asymmetric contact patterns when only a subset of fingers touches the surface. We use the same threshold $\tau_{\mathrm{inf}}=0.52$ on the mirrored predictions as for native right-hand inputs.

\end{document}